%% file: arxiv.tex
\documentclass{article}

\PassOptionsToPackage{numbers, compress}{natbib}

\usepackage[main, final]{neurips_2026}

\usepackage[utf8]{inputenc} 
\usepackage[T1]{fontenc}    
\usepackage{hyperref}       
\usepackage{url}            
\usepackage{booktabs}       
\usepackage{amsfonts}       
\usepackage{nicefrac}       
\usepackage{microtype}      
\usepackage{xcolor}         

\usepackage{graphicx}
\usepackage[font=small]{caption} 
\usepackage{amsmath}
\usepackage{amssymb}
\usepackage{wrapfig}
\usepackage{multirow}
\usepackage{multicol}
\usepackage[capitalize]{cleveref}

\DeclareMathOperator*{\argmax}{arg\,max}

\title{Early Semantic Grounding in Image Editing Models for Zero-Shot Referring Image Segmentation}

\author{%
  Jingxuan He\textsuperscript{1} \qquad
  Xiyu Wang\textsuperscript{1} \qquad
  Yunke Wang\textsuperscript{1} \qquad
  Mengyu Zheng\textsuperscript{2} \qquad
  Chang Xu\textsuperscript{1,}\thanks{Corresponding author.} \\[4pt]
  {\normalfont \textsuperscript{1}The University of Sydney \qquad \textsuperscript{2}Beijing Jiaotong University}
}

\begin{document}

\maketitle

\input{sections/0_abstract}
\input{sections/1_introduction}
\input{sections/2_related_work}
\input{sections/3_method}
\input{sections/4_experiments}
\input{sections/5_conclusion}

\clearpage
\renewcommand{\acksection}{\section*{Acknowledgments}}
\begin{ack}
This work was supported in part by the Australian Research Council under Projects DP240101848 and FT230100549.
\end{ack}

{
\small
\bibliographystyle{plainnat}
\bibliography{neurips_2026}
}

\clearpage
{
\appendix
\setcounter{section}{0}
\setcounter{figure}{0}
\setcounter{table}{0}
\setcounter{equation}{0}
\renewcommand{\thesection}{A\arabic{section}}
\renewcommand{\thefigure}{A\arabic{figure}}
\renewcommand{\thetable}{A\arabic{table}}
\renewcommand{\theequation}{A\arabic{equation}}
\input{sections/x_appendix}
}

\end{document}

%% file: sections/0_abstract.tex
\begin{abstract}
Instruction-based image editing (IIE) models have recently demonstrated strong capability in modifying specific image regions according to natural language instructions, which implicitly requires identifying where an edit should be applied. This indicates that such models inherently perform language-conditioned visual semantic grounding. In this work, we investigate whether this implicit grounding can be leveraged for zero-shot referring image segmentation (RIS), a task that requires pixel-level localization of objects described by natural language expressions. Through systematic analysis, we reveal that strong foreground-background separability is already present in the internal representations of these models at the earliest denoising timestep, well before any visible image transformation occurs. Building on this insight, we propose a training-free framework that repurposes pretrained image editing models for RIS by exploiting their intermediate representations. Our approach decomposes localization into two complementary components: attention-based spatial priors that estimate where to focus, and feature-based semantic discrimination that determines what to segment. By leveraging feature-space separability, the framework produces accurate segmentation masks using only a single denoising step, without requiring full image synthesis. Extensive experiments on RefCOCO, RefCOCO+, and RefCOCOg demonstrate that our method achieves superior performance over existing zero-shot baselines. Project page: \url{https://jessie459.github.io/EarlySemanticGrounding/}.
\end{abstract}

%% file: sections/1_introduction.tex
\section{Introduction}
\label{sec:introduction}

Instruction-based Image Editing (IIE) has recently witnessed significant progress.
By jointly modeling textual and visual signals, modern image editing frameworks~\cite{zhang2025enabling,xiao2025omnigen,wu2025omnigen2,flux-2-2025,liu2025step1x,wu2025qwen} demonstrate strong capability in modifying specific regions in accordance with natural language instructions.
This ability implicitly requires the model to identify \emph{where} the edit should be applied, indicating that image editing models inherently perform language-conditioned visual semantic grounding.
In this work, we investigate whether this implicit grounding can be directly extracted and repurposed for Referring Image Segmentation (RIS)~\cite{hu2016segmentation}, a task that requires pixel-level localization of objects described by natural language expressions.

Despite recent progress in fully and weakly supervised RIS, these methods~\cite{dai2025deris,yu2025latent,cheng2025weakmcn,chen2025dvin} rely on costly pixel-level annotations.
Zero-shot RIS provides a more practical alternative, yet existing frameworks typically adopt a multi-stage pipeline that combines the Segment Anything Model (SAM)~\cite{kirillov2023segment,carion2025sam} for mask proposal and the CLIP~\cite{radford2021learning} model for mask selection.
This design suffers from two fundamental limitations.
First, SAM is primarily geometry-driven and lacks an intrinsic vision-language alignment ability to interpret complex referring expressions.
Second, recent RIS methods, such as HybridGL~\cite{liu2025hybrid} and RefAM~\cite{kukleva2025refam}, employ intricate linguistic preprocessing to improve localization accuracy by decomposing referring expressions into noun phrases and explicitly modeling spatial relationships, which introduces additional complexity and is prone to error propagation.

In contrast, instruction-based image editing models provide a unified framework that jointly models language understanding and visual reasoning.
We therefore take a fundamentally different perspective by \emph{recasting zero-shot RIS as an image editing problem}.
However, this reformulation raises a critical question: \emph{in what form, and at what stage, does an image editing model establish the localization of the target object?}

Through systematic analysis of latent representations, we uncover a key and previously underexplored property of modern image editing models~\cite{flux-2-2025,liu2025step1x}.
We find that \emph{object-level semantics are encoded in the feature space of clean image tokens}, and more importantly, that \emph{strong foreground-background separability is already present at the earliest denoising timestep}, well before any visible image transformation occurs.
This suggests that semantic grounding is largely established before visible synthesis, and that subsequent denoising primarily refines visual appearance rather than object-level discrimination.
This insight also implies that localization signals can be extracted directly from early representations without requiring full image synthesis.

Building on these observations, we propose a training-free framework that repurposes pretrained image editing models for zero-shot referring image segmentation.
Rather than performing iterative denoising, which remains costly even with optimized sampling schedules~\cite{wang2023learning}, we extract internal representations from a single timestep and leverage them for localization.
Our approach decomposes the localization process into two complementary components.
First, we derive a coarse spatial prior from cross-attention, which captures instruction-conditioned relevance.
To improve spatial coherence, we introduce an affinity-based propagation mechanism that diffuses attention responses across semantically related regions.
Second, we exploit the strong semantic separability of latent features by constructing foreground and background prototypes, and perform pixel-wise classification via cosine similarity to obtain the final segmentation result.
This design explicitly decouples \emph{where} to look (attention) from \emph{what} to segment (feature semantics), leading to accurate and robust localization.
Extensive experiments on RefCOCO~\cite{nagaraja2016modeling}, RefCOCO+~\cite{nagaraja2016modeling}, and RefCOCOg~\cite{mao2016generation} demonstrate that our method consistently outperforms existing zero-shot baselines.

The contributions of this work are summarized as follows:
\begin{itemize}
    \item We reformulate referring image segmentation as an instruction-based image editing problem, showing that generative models inherently support discriminative visual grounding without requiring task-specific supervision.
    \item We reveal that modern image editing models exhibit strong foreground-background separability in their internal representations at the earliest denoising timestep.
    \item We propose a training-free framework that decomposes localization into two complementary components: attention-based spatial priors that estimate \emph{where} to focus, and feature-based semantic discrimination that determines \emph{what} to segment, resulting in precise and robust localization.
    \item Extensive experiments on RefCOCO, RefCOCO+, and RefCOCOg demonstrate that our method consistently outperforms existing zero-shot baselines, supported by comprehensive ablation studies that validate the effectiveness of each component.
\end{itemize}

%% file: sections/2_related_work.tex
\section{Related Work}

\noindent \textbf{Instruction-based Image Editing.}
Instruction-based image editing aims to modify images according to natural language instructions while preserving content outside the edited regions.
Early methods built upon Stable Diffusion~\cite{ho2020denoising,rombach2022high} learn to translate textual instructions into targeted visual modifications, as exemplified by InstructPix2Pix~\cite{brooks2023instructpix2pix}, MagicBrush~\cite{zhang2023magicbrush}, UltraEdit~\cite{zhao2024ultraedit}, and AnyEdit~\cite{yu2025anyedit}.
More recent approaches, such as ICEdit~\cite{zhang2025enabling}, improve synthesis fidelity by fine-tuning stronger Diffusion Transformer (DiT) backbones~\cite{peebles2023scalable}.
To further enhance instruction following, several state-of-the-art frameworks~\cite{liu2025step1x,wu2025qwen,cai2025z,team2025longcat} incorporate multi-modal large language models to strengthen cross-modal alignment between textual and visual representations.
While these models are primarily designed for generative tasks, their ability to interpret editing instructions implies an inherent capacity for visual grounding.
Recent work further shows that edit regions can be localized inside DiT-based editors from attention-guided feature centroids, which improves content consistency outside the edited regions~\cite{he2026rethinking}.
Building on this intuition, we repurpose instruction-based image editing models for referring image segmentation in a zero-shot manner.

\noindent \textbf{Referring Image Segmentation.}
Referring image segmentation (RIS) is a visual grounding task that aims to segment objects specified by natural language expressions, which requires models to establish fine-grained associations between visual content and linguistic cues.
While fully supervised~\cite{chng2024mask,liu2024rotated,shah2024lqmformer,shang2024prompt,dai2025deris,yu2025latent} and weakly supervised~\cite{dai2024curriculum,yang2024boosting,cheng2025weakmcn,chen2025dvin} RIS methods have achieved strong performance, they remain heavily reliant on dense pixel-level annotations.
Zero-shot referring image segmentation~\cite{yu2023zero,ni2023ref,suo2023text,sun2024clip,wang2025iterprime,liu2025hybrid,kukleva2025refam} offers a more scalable alternative by leveraging pretrained foundation models.
Global-Local~\cite{yu2023zero} is an early representative that exploits pretrained FreeSOLO~\cite{wang2022freesolo} and CLIP~\cite{radford2021learning} for mask prediction.
CaR~\cite{sun2024clip} improves upon this by iteratively refining masks through a CLIP-based recurrent framework.
TAS~\cite{suo2023text} enhances segmentation performance with descriptive captions generated by an auxiliary captioner.
More recently, HybridGL~\cite{liu2025hybrid} introduces a hybrid global-local feature extraction strategy with spatial guidance to mitigate segmentation ambiguity, and RefAM~\cite{kukleva2025refam} augments referring expressions with stop words to suppress attention sinks.
Despite these advances, most existing zero-shot RIS methods remain dependent on SAM~\cite{kirillov2023segment} for mask proposals and thus inherit its class-agnostic limitations in resolving referring expressions.
In contrast, our framework directly harnesses the semantic grounding capability of pretrained image editing models for referring image segmentation.


\noindent \textbf{Segmentation with Attention and Diffusion Features.}
Attention and affinity have been widely used to capture spatial relationships for segmentation.
Ru et al.~\cite{ru2022learning} and Xu et al.~\cite{xu2022multi} learn affinities from transformer attention for weakly supervised segmentation, while DiffSeg~\cite{tian2024diffuse} and DiffCut~\cite{couairon2024diffcut} exploit self-attention maps and diffusion features, respectively, for unsupervised segmentation.
More recently, ConceptAttention~\cite{helbling2025conceptattention} localizes concepts in text-to-image MM-DiTs through similarities between image representations and separately introduced concept representations.
RefAM~\cite{kukleva2025refam} uses DiT cross-attention to localize referring expressions and obtains segmentation masks with SAM.
Our work instead examines clean reference-image tokens in instruction-based editing models and shows that they already exhibit strong foreground-background separability at the earliest evaluated denoising step.
Building on this finding, our training-free framework uses attention-derived spatial priors to construct foreground and background prototypes, then predicts referring masks through feature-space classification in a single denoising step.

%% file: sections/3_method.tex
\section{Method}
\label{sec:method}

\subsection{Preliminaries: Instruction-based Image Editing}
\label{sec:preliminaries}

Recent multi-modal image editing models~\cite{flux2024,flux-2-2025,liu2025step1x,yin2025reasonedit} typically adopt a Large Language Model (LLM)~\cite{bai2025qwen2} to encode editing instructions and a Multi-Modal Diffusion Transformer (MM-DiT)~\cite{peebles2023scalable,esser2024scaling} as a unified generator that jointly models textual and visual signals.
Given an editing instruction $P$ and a reference image $I \in \mathbb{R}^{H \times W \times 3}$, the LLM encodes $P$ (and $I$ in multi-modal settings) into a sequence of prompt embeddings $\mathbf{Z}_p \in \mathbb{R}^{N_p \times D}$, where $H$ and $W$ denote the height and width of the reference image, $N_p$ is the number of prompt tokens, and $D$ is the embedding dimension.
For notational simplicity, we use ``tokens'' to refer to token embeddings throughout this work.
The MM-DiT operates on three types of tokens: prompt tokens, clean image tokens $\mathbf{Z}_v \in \mathbb{R}^{N_v \times D}$, and noisy image tokens $\hat{\mathbf{Z}}_v^{(t)} \in \mathbb{R}^{N_v \times D}$ at diffusion timestep $t$, where $N_v$ is the number of image tokens.
These tokens are concatenated along the sequence dimension~\cite{tan2025ominicontrol} and passed through a stack of transformer blocks, where the Multi-Modal Self-Attention (MMSA) module enables information exchange across modalities.
Specifically, we denote the input to MMSA within the $l$-th DiT block at timestep $t$ as $\mathbf{Z}^{(t,l)} = [ \mathbf{Z}_{p}^{(l)}; \hat{\mathbf{Z}}_{v}^{(t,l)}; \mathbf{Z}_{v}^{(l)} ] \in \mathbb{R}^{( N_p + 2N_v) \times D}$.
This unified token sequence $\mathbf{Z}^{(t,l)}$ is then transformed into queries $\mathbf{Q}^{(t,l)}$, keys $\mathbf{K}^{(t,l)}$, and values $\mathbf{V}^{(t,l)}$ through linear projection.
The MMSA module is formulated as:
\begin{equation}
\label{eq:self_attention}
    \text{MMSA}\left ( \mathbf{Z}^{(t,l)} \right ) = \mathbf{A}^{(t,l)} \mathbf{V}^{(t,l)},
    \; \text{where} \;
    \mathbf{A}^{(t,l)} = \operatorname{Softmax} \left( \frac{\mathbf{Q}^{(t,l)} {\mathbf{K}^{(t,l)}}^{\top}}{\sqrt{D}} \right). 
\end{equation}
We denote the full attention matrix as $\mathbf{A}^{(t,l)} \in \mathbb{R}^{(N_p + 2N_v) \times (N_p + 2N_v)}$.
Starting from a Gaussian noise, the noisy image tokens $\hat{\mathbf{Z}}_v^{(t)}$ are iteratively denoised across diffusion timesteps, and the fully denoised tokens are decoded to produce the final edited image.

\subsection{Referring Image Segmentation}

\begin{figure}[t]
    \small
    \centering
    \includegraphics[width=\linewidth]{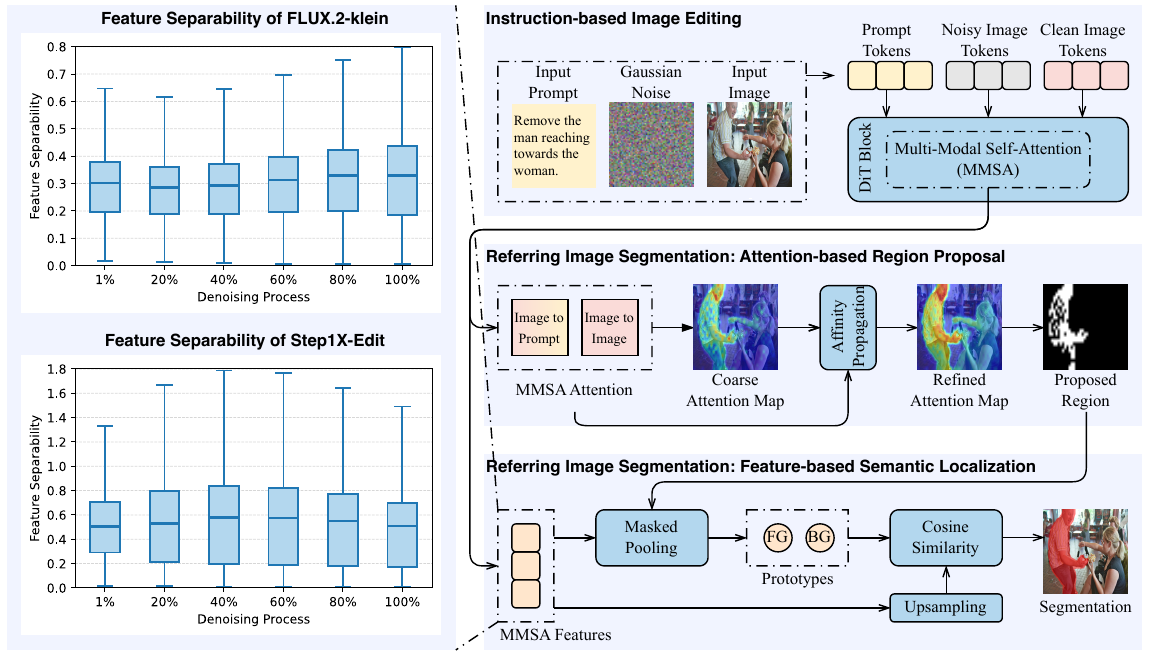}
    \caption{\textbf{Overview of our framework.} We repurpose pretrained instruction-based image editing models for referring image segmentation. In the module of attention-based region proposal, we derive coarse spatial priors through cross-attention and affinity propagation. In the module of feature-based semantic localization, the semantic separability of MMSA output features is exploited under the guidance of the proposed region to produce the final segmentation mask.
    The quantitative results in boxplots on the left panel show that strong foreground-background discriminability is already present at the earliest evaluated timestep (denoted as 1\%) and remains relatively stable throughout the denoising process.
    }
    \label{fig:overview}
\end{figure}

Given an input image $I$ and a referring expression $T_{\text{ref}}$, the goal of referring image segmentation is to produce a binary mask that identifies the referred object in the image.
We reformulate this task as an instance of instruction-based image editing, where the referring expression $T_{\text{ref}}$ is converted into an editing instruction $T$ by prepending ``remove'' to the referring expression.
We further refine the instructions using the built-in instruction-refinement pipeline of Step1X-Edit~\cite{yin2025reasonedit}, which preserves the original meaning and retains instructions that are already clear.
Details and the effect of refinement are provided in Appendix~\ref{sec:instruction_refinement}, and the robustness to alternative instruction templates is analyzed in \cref{sec:ablation}.

Although image editing models are primarily designed for image synthesis, their denoising process inherently requires identifying regions relevant to the editing instruction.
This suggests that intermediate representations may already encode object-level localization signals.
Motivated by this insight, we first analyze the foreground-background separability of MMSA features extracted from clean image tokens, and then develop a two-stage framework that combines attention-based spatial priors with feature-based semantic discrimination for precise localization.

\noindent \textbf{Feature Separability Analysis.}
Let $\mathbf{F}^{(t,l)} \in \mathbb{R}^{N_v \times D}$ denote MMSA output features of clean image tokens within the $l$-th DiT block at timestep $t$, and let $\Omega_{\text{fg}}$ and $\Omega_{\text{bg}}$ denote the sets of foreground and background spatial locations, respectively.
We compute class-wise means $\mu_{\text{fg}}^{(t,l)}, \mu_{\text{bg}}^{(t,l)}$ and variances $\sigma_{\text{fg}}^{2(t,l)}, \sigma_{\text{bg}}^{2(t,l)}$ from these features.
Inspired by the Fisher discriminant criterion~\cite{fisher1936use}, we define the separability score as:
\begin{equation}
    \mathcal{J}^{(t,l)} = \frac{\left\| \mu_{\text{fg}}^{(t,l)} - \mu_{\text{bg}}^{(t,l)} \right\|^{2}}{\sigma_{\text{fg}}^{2(t,l)} + \sigma_{\text{bg}}^{2(t,l)}}.
\end{equation}
This metric measures the ratio between inter-class distance and intra-class variance, where larger values indicate stronger feature discriminability.
Detailed equations and rationale for this metric are provided in Appendix~\ref{app:feature_separability}.

We evaluate $\mathcal{J}^{(t,l)}$ on FLUX.2-klein~\cite{flux-2-2025} and Step1X-Edit~\cite{liu2025step1x} across a range of timesteps $t$.
The DiT block $l$ is set to approximately two-thirds of the model depth.
As illustrated in the left panel of \cref{fig:overview}, both models exhibit strong feature separability as early as the first denoising timestep, and the separability remains relatively stable throughout the denoising process.
Quantitative results across three benchmarks further support this observation (\cref{tab:separability}).
This indicates that \emph{semantic grounding is already established at early timesteps}, while subsequent timesteps may focus on fine-grained visual synthesis rather than improving object-level discrimination.

\noindent \textbf{Attention-based Region Proposal (ARP).}
Although MMSA features exhibit strong discriminative capability at early timesteps, they do not explicitly capture the spatial extent of the referred object.
We therefore first derive a coarse estimate of the target region from attention, which serves as a spatial prior for subsequent feature-based localization.

As shown in the top right panel of \cref{fig:overview}, the $l$-th DiT block takes as input the prompt tokens $\mathbf{Z}_{p}^{(l)}$, the clean image tokens $\mathbf{Z}_{v}^{(l)}$, and the noisy image tokens $\hat{\mathbf{Z}}_{v}^{(t,l)}$ at timestep $t$.
After token-wise concatenation, the MMSA module processes the unified token sequence by projecting it into queries $\mathbf{Q}^{(t,l)}$, keys $\mathbf{K}^{(t,l)}$, and values $\mathbf{V}^{(t,l)}$, respectively, followed by the standard self-attention operation defined in \cref{eq:self_attention}.
Next, as illustrated in the middle right panel of \cref{fig:overview}, we extract the image-to-prompt submatrix $\mathbf{A}_{v \rightarrow p}^{(t,l)} \in \mathbb{R}^{N_v \times N_p}$ from the full attention matrix.
This submatrix is aggregated across $N_p$ prompt tokens and $N_l$ DiT blocks, resulting in the Coarse Attention Map (CAM) $\mathbf{A}_{\text{CAM}}^{(t)} \in \mathbb{R}^{N_h \times N_w}$:
\begin{equation}
    \mathbf{A}_{\text{CAM}}^{(t)} = \mathcal{N} \left( \sum_{l=1}^{N_l} \mathcal{F}^{-1} \left( \mathbf{A}_{v \rightarrow p}^{(t,l)} \mathbf{1}_{N_p} \right) \right),
\end{equation}
where $\mathbf{1}_{N_p}$ denotes an all-ones vector of length $N_p$, $\mathcal{F}^{-1}(\cdot)$ reshapes a flattened sequence of length $N_v$ into a spatial map of size $N_h \times N_w$, and $\mathcal{N}(\cdot)$ denotes min-max normalization.

Although the coarse attention map identifies an approximate location of the referred object, it often exhibits fragmented or noisy activations due to the lack of spatial coherence in cross-attention.
To alleviate this issue, inspired by \cite{xu2022multi,ru2022learning}, we introduce an affinity-based propagation mechanism that leverages self-attention to propagate cross-attention responses across semantically related regions.
Specifically, we extract the image-to-image submatrix $\mathbf{A}_{v \rightarrow v}^{(t,l)} \in \mathbb{R}^{N_v \times N_v}$ from the full attention matrix, and interpret it as an affinity matrix that captures pairwise relationships among clean image tokens.
We perform row-wise normalization to obtain the stochastic transition matrix $\mathbf{T}^{(t,l)} \in \mathbb{R}^{N_v \times N_v}$, which is used to compute the Refined Attention Map (RAM) $\mathbf{A}_{\text{RAM}}^{(t)} \in \mathbb{R}^{N_h \times N_w}$:
\begin{equation}
    \mathbf{A}_{\text{RAM}}^{(t)} = \mathcal{N} \left( \sum_{l=1}^{N_l} \mathcal{F}^{-1} \left( \mathbf{T}^{(t,l)} \mathbf{A}_{v \rightarrow p}^{(t,l)} \mathbf{1}_{N_p} \right) \right),
    \: \text{where} \:
    \mathbf{T}^{(t,l)} = \mathcal{D} \left( \mathbf{A}_{v \rightarrow v}^{(t,l)} \mathbf{1}_{N_v} \right)^{-1} \mathbf{A}_{v \rightarrow v}^{(t,l)}.
\end{equation}
Here $\mathcal{D}(\cdot)$ constructs a diagonal matrix from its input.
This attention propagation design can be interpreted as a random walk~\cite{lovasz1993random} over image tokens, where the stochastic transition matrix defines a Markov process based on pairwise affinities.
Since each affinity is determined by the query-key similarity between two clean-image tokens, tokens within the same semantic region, which share appearance and context, generally exhibit stronger affinities.
Attention is therefore diffused to semantically coherent regions, resulting in improved spatial continuity and better alignment with object boundaries.

\noindent \textbf{Feature-based Semantic Localization (FSL).}
While attention-based region proposal provides necessary spatial priors, it remains inherently imprecise due to the diffuse nature of attention.
In contrast, our early analysis shows that MMSA features exhibit strong foreground-background separability, suggesting that object-level semantics are already well captured in the feature space.
We therefore leverage the rich semantics of these features to transform the initial proposal into an accurate localization of the referred object.

As depicted in the bottom right panel of \cref{fig:overview}, we extract MMSA features of clean image tokens at timestep $t^{\prime}$ and DiT block $l^{\prime}$.
These features are reshaped into a spatial feature map and $\ell_2$-normalized to lie on the unit hypersphere, yielding $\hat{\mathbf{F}}^{(t^{\prime},l^{\prime})} \in \mathbb{R}^{D \times N_h \times N_w}$. 
We then obtain a binary mask $\mathbf{M}^{(t)} = \mathbf{1}[\mathbf{A}_{\text{RAM}}^{(t)} > \tau]$ by thresholding the refined attention map, which selects high-confidence regions of the referred object for prototype construction, where $\tau$ is a predefined threshold and $\mathbf{1}[\cdot]$ denotes the indicator function.
Foreground and background semantic prototypes, $\mathbf{P}_{\mathrm{fg}} \in \mathbb{R}^{D}$ and $\mathbf{P}_{\mathrm{bg}} \in \mathbb{R}^{D}$, are constructed via masked average pooling over the normalized features:
\begin{equation}
    \begin{aligned}
        \mathbf{P}_{\mathrm{fg}} &=
        \frac{\sum_{h=1}^{N_h}\sum_{w=1}^{N_w}
        \mathbf{M}^{(t)}_{h,w}\hat{\mathbf{F}}^{(t^{\prime},l^{\prime})}_{:,h,w}}
        {\sum_{h=1}^{N_h}\sum_{w=1}^{N_w}\mathbf{M}^{(t)}_{h,w}},\\
        \mathbf{P}_{\mathrm{bg}} &=
        \frac{\sum_{h=1}^{N_h}\sum_{w=1}^{N_w}
        \left(1-\mathbf{M}^{(t)}_{h,w}\right)\hat{\mathbf{F}}^{(t^{\prime},l^{\prime})}_{:,h,w}}
        {\sum_{h=1}^{N_h}\sum_{w=1}^{N_w}\left(1-\mathbf{M}^{(t)}_{h,w}\right)}.
    \end{aligned}
\end{equation}
The feature map $\hat{\mathbf{F}}^{(t^{\prime},l^{\prime})}$ is subsequently upsampled to the original image resolution, yielding $\hat{\mathbf{F}}_{\mathrm{up}}^{(t^{\prime},l^{\prime})} \in \mathbb{R}^{D \times H \times W}$.
We perform pixel-wise binary classification by computing cosine similarity between each prototype and the upsampled features.
Each pixel is assigned to the class whose prototype has higher similarity. 
The final segmentation mask $\mathbf{S} \in \{0,1\}^{H \times W}$ is obtained as:
\begin{equation}
    \mathbf{S}_{h,w} =
    \argmax_{c \in \{\mathrm{fg},\mathrm{bg}\}}
    \mathbf{P}_{c}^{\top}
    \hat{\mathbf{F}}_{\mathrm{up},:,h,w}^{(t^{\prime},l^{\prime})}.
\end{equation}
We empirically find that performing attention estimation at $t=0$ and feature extraction at $t^{\prime}=0$ within a deep DiT block yields the best performance on the validation sets (\cref{sec:ablation}).
This configuration provides both an accurate and efficient solution for referring image segmentation.

%% file: sections/4_experiments.tex
\section{Experiments}
\label{sec:experiments}

\subsection{Datasets and Metrics}
Following previous works~\cite{suo2023text,wang2025iterprime,liu2025hybrid,kukleva2025refam}, we conduct experiments on three standard Referring Image Segmentation (RIS) benchmarks: RefCOCO~\cite{nagaraja2016modeling}, RefCOCO+~\cite{nagaraja2016modeling}, and RefCOCOg~\cite{mao2016generation}.
All images in these benchmarks are drawn from the MSCOCO dataset~\cite{lin2014microsoft} and are paired with carefully annotated referring expressions that specify target objects.
RefCOCO contains relatively concise expressions, many of which include positional cues such as ``left'' and ``right'', while such spatial terms are excluded in RefCOCO+. 
RefCOCOg is a more challenging benchmark that features longer and more complex expressions.
We adopt two commonly used metrics to evaluate segmentation performance: overall Intersection over Union (oIoU) and mean Intersection over Union (mIoU).
Specifically, oIoU computes intersections and unions over all samples before taking their ratio, thereby emphasizing performance on larger objects, while mIoU averages IoU scores across individual samples and reflects performance across varying object scales.

\subsection{Implementation Details}
We repurpose two state-of-the-art instruction-based image editing models, FLUX.2-klein~\cite{flux-2-2025} and Step1X-Edit~\cite{liu2025step1x}, for zero-shot referring image segmentation.
To obtain Coarse Attention Maps (CAMs) and Refined Attention Maps (RAMs), we perform a single denoising step at $t=0$ and aggregate attention weights across all DiT blocks.
For feature-based semantic localization, we extract MMSA output features at $t^{\prime}=0$.
Due to differences in model depth, we select features from $l^{\prime}=20$ out of 32 DiT blocks for FLUX.2-klein and $l^{\prime}=40$ out of 57 DiT blocks for Step1X-Edit.
All hyperparameters ($t=t^{\prime}=0$, $l^{\prime}=40$, and $\tau=0.8$) are selected once using Step1X-Edit on the \textit{val} splits (\cref{sec:ablation}) and fixed for all datasets and test splits; for FLUX.2-klein, we transfer the selected relative depth (about two-thirds) to its 32 blocks.
All experiments are conducted on one NVIDIA A100 GPU.
A detailed analysis of inference efficiency, including latency and memory comparisons with baselines, is provided in Appendix~\ref{app:inference_efficiency}.

\begin{table*}[t]
    \centering
    \small
    \caption{\textbf{Comparisons with state-of-the-art zero-shot RIS methods on RefCOCO, RefCOCO+, and RefCOCOg}. The best and second-best results are highlighted in \textbf{bold} and \underline{underlined}, respectively. The notation of $^*$ indicates the use of additional data beyond the task-specific dataset. The number in parentheses to the right of the model name denotes the parameter count of the corresponding DiT model.}
    \label{tab:result}
    \setlength{\tabcolsep}{4pt}
    \resizebox{\textwidth}{!}{\begin{tabular}{c|lll|ccc|ccc|cc}
    \toprule
    \multirow{2}{*}{Metric} & \multirow{2}{*}{Method} & \multirow{2}{*}{Vision Backbone} & \multirow{2}{*}{Pretrained Model} & \multicolumn{3}{c|}{RefCOCO} & \multicolumn{3}{c|}{RefCOCO+} & \multicolumn{2}{c}{RefCOCOg} \\
    & & & & val & testA & testB & val & testA & testB & val & test \\
    \midrule
    \multirow{16}{*}{oIoU}
    & \multicolumn{3}{l|}{\textit{zero-shot methods w/ additional training}} &  &  &  &  &  &  &  & \\
    & {Pseudo-RIS}~\cite{yu2024pseudo} & ViT-B  & SAM, CoCa, CLIP          & 37.33 & 43.43 & 31.90 & 40.19 & 46.43 & 33.63 & 41.63 & 43.52 \\
    & {VLM-VG}~\cite{wang2025learning} & R101   & COCO$^{*}$, VLM-VG$^{*}$ & 45.40 & 48.00 & 41.40 & 37.00 & 40.70 & 30.50 & 42.80 & 44.10 \\
    \cline{2-12}
    & \multicolumn{3}{l|}{\textit{zero-shot methods w/o additional training}} & & & & & & & & \\
    & {Grad-CAM}~\cite{selvaraju2017grad} & R50   & SAM, CLIP        & 23.44 & 23.91 & 21.60 & 26.67 & 27.20 & 24.84 & 23.00 & 23.91 \\
    & {MaskCLIP}~\cite{zhou2022extract}   & R50   & SAM, CLIP        & 20.18 & 20.52 & 21.30 & 22.06 & 22.43 & 24.61 & 23.05 & 23.41 \\
    & {Global-Local}~\cite{yu2023zero}    & R50   & FreeSOLO, CLIP   & 24.58 & 23.38 & 24.35 & 25.87 & 24.61 & 25.61 & 30.07 & 29.83 \\
    & {Global-Local}~\cite{yu2023zero}    & R50   & SAM, CLIP        & 24.55 & 26.00 & 21.03 & 26.62 & 29.99 & 22.23 & 28.92 & 30.48 \\
    & {Global-Local}~\cite{yu2023zero}    & ViT-B & SAM, CLIP        & 21.71 & 24.48 & 20.51 & 23.70 & 28.12 & 21.86 & 26.57 & 28.21 \\
    & {Ref-Diff}~\cite{ni2023ref}         & ViT-B & SAM, SD, CLIP    & 35.16 & 37.44 & 34.50 & 35.56 & 38.66 & 31.40 & 38.62 & 37.50 \\
    & {TAS}~\cite{suo2023text}            & ViT-B & SAM, BLIP2, CLIP & 29.53 & 30.26 & 28.24 & 33.21 & 38.77 & 28.01 & 35.84 & 36.16 \\
    & {SAM 3}~\cite{carion2025sam}        & PE    & SAM 3            & 38.77 & 37.43 & 41.80 & 31.62 & 32.44 & 30.70 & 35.04 & 35.81 \\
    & {HybridGL}~\cite{liu2025hybrid}     & ViT-B & SAM, CLIP        & 41.81 & 44.52 & 38.50 & 35.74 & 41.43 & 30.90 & 42.47 & 42.97 \\
    & {RefAM}~\cite{kukleva2025refam}     & DiT   & SAM, FLUX.1-dev (12B) & 46.91 & 52.30 & 43.88 & 38.57 & 42.66 & 34.90 & 45.53 & 44.45 \\
    & Ours                                & DiT   & FLUX.2-klein (9B)& \underline{50.13} & \underline{52.99} & \underline{46.10} & \underline{40.17} & \underline{43.25} & \underline{36.06} & \underline{46.90} & \underline{46.26} \\
    & Ours                                & DiT   & Step1X-Edit (12B)& \textbf{53.94}    & \textbf{58.10}    & \textbf{49.60}    & \textbf{44.38}    & \textbf{49.30}    & \textbf{40.06}    & \textbf{51.23}    & \textbf{50.86} \\
    \hline
    \hline
    \multirow{17}{*}{mIoU}
    & \multicolumn{3}{l|}{\textit{zero-shot methods w/ additional training}} & & & & & & & & \\
    & {Pseudo-RIS}~\cite{yu2024pseudo} & ViT-B & SAM, CoCa, CLIP          & 41.05 & 48.19 & 33.48 & 44.33 & 51.42 & 35.08 & 45.99 & 46.67 \\
    & {VLM-VG}~\cite{wang2025learning} & R101  & COCO$^{*}$, VLM-VG$^{*}$ & 49.90 & 53.10 & 46.70 & 42.70 & 47.30 & 36.20 & 48.00 & 48.50 \\
    \cline{2-12}
    & \multicolumn{3}{l|}{\textit{zero-shot methods w/o additional training}} & & & & & & & & \\
    & {Grad-CAM}~\cite{selvaraju2017grad} & R50   & SAM, CLIP        & 30.22 & 31.90 & 27.17 & 33.96 & 25.66 & 32.29 & 33.05 & 32.50 \\
    & {MaskCLIP}~\cite{zhou2022extract}   & R50   & SAM, CLIP        & 25.62 & 26.66 & 25.17 & 27.49 & 28.49 & 30.47 & 30.13 & 30.15 \\
    & {Global-Local}~\cite{yu2023zero}    & R50   & FreeSOLO, CLIP   & 26.70 & 24.99 & 26.48 & 28.22 & 26.54 & 27.86 & 33.02 & 33.12 \\
    & {Global-Local}~\cite{yu2023zero}    & R50   & SAM, CLIP        & 31.83 & 32.93 & 28.64 & 34.97 & 37.11 & 30.61 & 40.66 & 40.94 \\
    & {Global-Local}~\cite{yu2023zero}    & ViT-B & SAM, CLIP        & 33.12 & 36.52 & 29.58 & 35.29 & 39.58 & 31.89 & 40.08 & 40.74 \\
    & {Ref-Diff}~\cite{ni2023ref}         & ViT-B & SAM, SD, CLIP    & 37.21 & 38.40 & 37.19 & 37.29 & 40.51 & 33.01 & 44.02 & 44.51 \\
    & {TAS}~\cite{suo2023text}            & ViT-B & SAM, BLIP2, CLIP & 39.84 & 41.08 & 36.24 & 43.63 & \underline{49.13} & 36.54 & 46.62 & 46.80 \\
    & {CaR}~\cite{sun2024clip}            & ViT-B and ViT-L & CLIP   & 33.57 & 35.36 & 30.51 & 34.22 & 36.03 & 31.02 & 36.67 & 36.57 \\
    & {SAM 3}~\cite{carion2025sam}        & PE    & SAM 3            & 34.80 & 33.36 & 38.91 & 26.83 & 28.30 & 25.56 & 32.14 & 32.08 \\
    & {HybridGL}~\cite{liu2025hybrid}     & ViT-B & SAM, CLIP        & 49.48 & 53.37 & 45.19 & 43.40 & \underline{49.13} & 37.17 & \underline{51.25} & \underline{51.59} \\
    & {RefAM}~\cite{kukleva2025refam}     & DiT   & SAM, FLUX.1-dev (12B) & \underline{57.24} & \underline{59.78} & \underline{53.32} & 43.59             & 47.28          & 38.77             & 47.11          & 48.35 \\
    & Ours                             & DiT   & FLUX.2-klein (9B)   & 54.02             & 56.09             & 50.22             & \underline{44.15} & 46.72          & \underline{40.25} & 50.62          & 50.47 \\
    & Ours                             & DiT   & Step1X-Edit (12B)   & \textbf{57.56}    & \textbf{60.34}    & \textbf{53.76}    & \textbf{48.50}    & \textbf{52.37} & \textbf{44.28}    & \textbf{54.52} & \textbf{54.25} \\
    \bottomrule
\end{tabular}}
\end{table*}

\begin{figure*}[t]
    \centering
    \includegraphics[width=\linewidth]{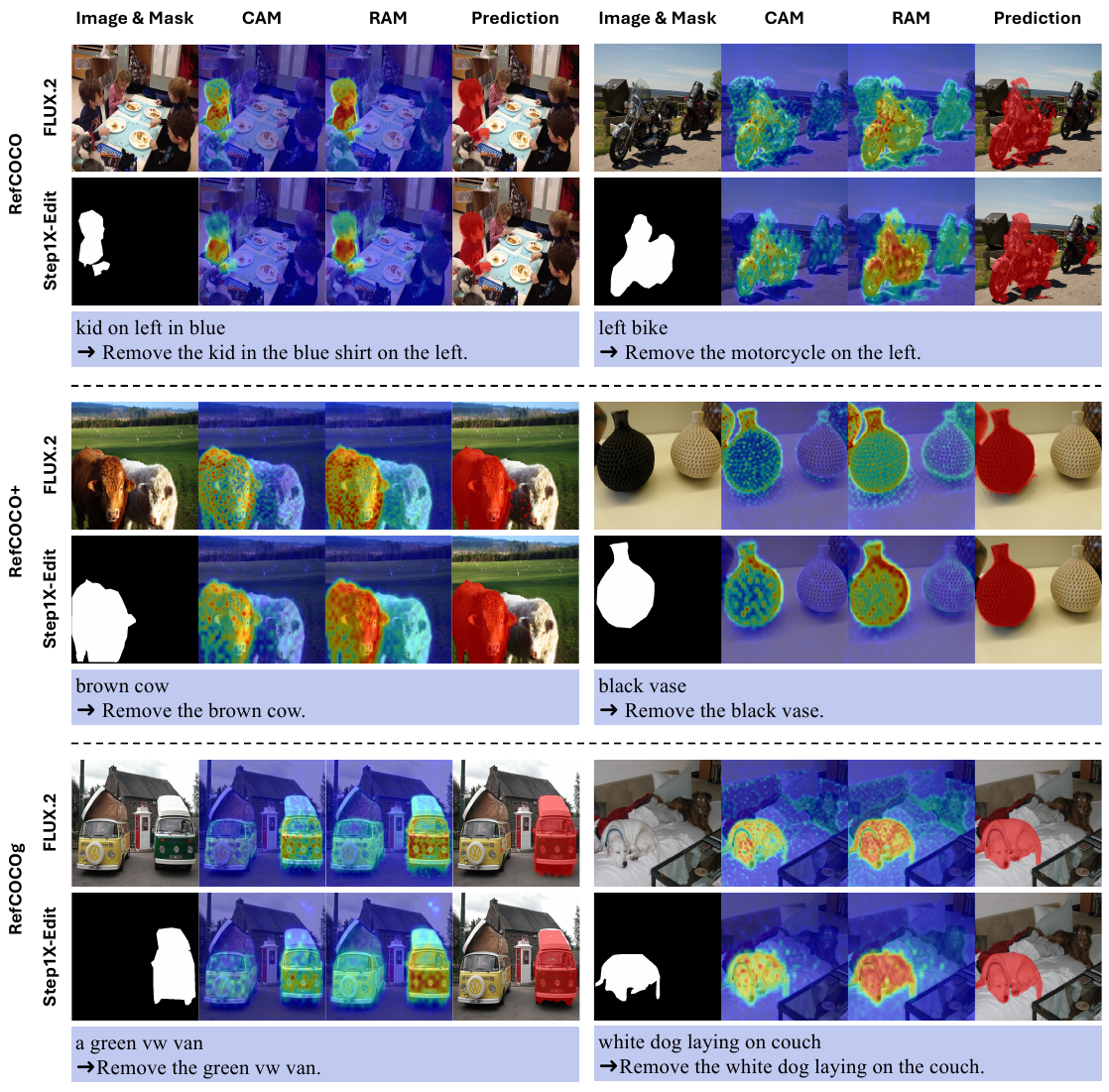}
    \caption{\textbf{Qualitative examples from RefCOCO, RefCOCO+, and RefCOCOg datasets.} ``CAM'' denotes Coarse Attention Map, ``RAM'' denotes Refined Attention Map, and ``GT'' denotes Ground Truth. The first row of text presents the original referring expressions, and the second row presents the corresponding transformed editing instructions.}
    \label{fig:ablation_modules}
\end{figure*}

\subsection{Comparison with State-of-the-arts}
\cref{tab:result} presents quantitative comparisons with state-of-the-art zero-shot referring image segmentation methods on RefCOCO, RefCOCO+, and RefCOCOg.
Overall, both FLUX.2-klein and Step1X-Edit variants achieve competitive performance across all three benchmarks, with Step1X-Edit consistently attaining the best results across dataset splits under both oIoU and mIoU metrics.
In terms of oIoU, the Step1X-Edit variant surpasses the strongest baseline RefAM~\cite{kukleva2025refam}, which uses a DiT of the same parameter scale (12B), by a clear margin of approximately 5-7 points, while the FLUX.2-klein (9B) variant consistently achieves the second-best results.
For mIoU, the Step1X-Edit variant outperforms existing approaches across all three benchmarks, indicating more robust segmentation across objects of varying scales.
Notably, although our framework requires no training, the Step1X-Edit variant also outperforms Pseudo-RIS~\cite{yu2024pseudo} and VLM-VG~\cite{wang2025learning}, which rely on additional training, on all dataset splits under both metrics.
Unlike prior methods that rely on explicit positional modeling or directional bias (e.g., Ref-Diff~\cite{ni2023ref}, MaskCLIP~\cite{zhou2022extract}, Global-Local~\cite{yu2023zero}, HybridGL~\cite{liu2025hybrid}, RefAM~\cite{kukleva2025refam}), our approach directly leverages the semantic grounding capability of pretrained image editing models.
This also explains the relatively smaller mIoU gains of our method on RefCOCO, where expressions frequently depend on directional cues.
Although SAM 3~\cite{carion2025sam} is specifically designed for segmentation, it performs suboptimally on the RIS task due to its class-agnostic nature.
In particular, it often misses the referred object entirely or segments all instances of the same category, regardless of the spatial relations specified in the expression.
Additional analysis and qualitative examples of SAM 3 are provided in Appendix~\ref{app:sam3}.

\begin{table}[t]
    \centering
    \begin{minipage}{0.52\textwidth}
        \centering
        \small
        \setlength{\tabcolsep}{4pt}
        \caption{\textbf{Ablation of modules within our framework}. ``CAMs'', ``RAMs'', and ``FSL'' denote coarse attention maps, refined attention maps, and feature-based semantic localization, respectively.}
        \label{tab:ablation_modules}
        \resizebox{\textwidth}{!}{%
        \begin{tabular}{c|ccc|ccc}
        \toprule
        & CAMs & RAMs & FSL & RefCOCO & RefCOCO+ & RefCOCOg \\
        \midrule
        \multirow{3}{*}{oIoU} & \checkmark &            &            & 33.02 & 26.83 & 32.82 \\
                              & \checkmark & \checkmark &            & 49.56 & 40.23 & 47.24 \\
                              & \checkmark & \checkmark & \checkmark & \textbf{53.94} & \textbf{44.38} & \textbf{51.23} \\
        \hline
        \hline
        \multirow{3}{*}{mIoU} & \checkmark &            &            & 39.44 & 32.03 & 40.11 \\
                              & \checkmark & \checkmark &            & 53.36 & 43.82 & 51.04 \\
                              & \checkmark & \checkmark & \checkmark & \textbf{57.56} & \textbf{48.50} & \textbf{54.52} \\
        \bottomrule
        \end{tabular}}
    \end{minipage}
    \hfill
    \begin{minipage}{0.46\textwidth}
        \centering
        \small
        \setlength{\tabcolsep}{4pt}
        \caption{\textbf{Ablation of feature extraction location}. ``Block'', ``AdaLN'', and ``MMSA'' denote output features of a diffusion transformer block, adaptive layer normalization, and multi-modal self-attention, respectively.}
        \label{tab:ablation_feat_extraction}
        \resizebox{\textwidth}{!}{%
        \begin{tabular}{l|cc|cc|cc}
        \toprule
        \multirow{2}{*}{Feature} & \multicolumn{2}{c|}{RefCOCO} & \multicolumn{2}{c|}{RefCOCO+} & \multicolumn{2}{c}{RefCOCOg} \\
        & oIoU & mIoU & oIoU & mIoU & oIoU & mIoU \\
        \midrule
        Block & 52.07 & 56.67 & 41.25 & 45.83 & 47.93 & 53.25 \\
        AdaLN & 53.63 & 57.34 & 43.24 & 47.42 & 49.89 & 53.82 \\
        MMSA  & \textbf{53.94} & \textbf{57.56} & \textbf{44.38} & \textbf{48.50} & \textbf{51.23} & \textbf{54.52} \\
        \bottomrule
        \end{tabular}}
    \end{minipage}
\end{table}

\subsection{Ablation Studies}
\label{sec:ablation}

To assess the contribution of each module in our framework and the impact of hyperparameters, we conduct comprehensive ablation studies using the Step1X-Edit variant on the \textit{val} splits of RefCOCO, RefCOCO+, and RefCOCOg.

\noindent \textbf{Quantitative ablation of modules.}
We evaluate the contributions of Attention-based Region Proposal (ARP) and Feature-based Semantic Localization (FSL).
For ablation experiments, attention maps are binarized into segmentation masks using a fixed threshold of 0.5.
As shown in \cref{tab:ablation_modules}, CAMs alone yield limited performance, indicating that coarse spatial priors are insufficient for accurate segmentation.
Introducing RAMs leads to substantial improvements across all datasets (e.g., around 16-point oIoU gain on RefCOCO).
This shows that affinity propagation effectively enhances spatial coherence.
Incorporating FSL further improves oIoU and mIoU, which highlights the importance of leveraging semantically separable latent features for precise foreground-background discrimination.
Overall, these results demonstrate that accurate segmentation arises from the complementary roles of attention-based spatial priors and feature-based semantic localization.

\noindent \textbf{Qualitative ablation of modules.}
We present qualitative examples in \cref{fig:ablation_modules}.
As illustrated, CAMs provide initial but spatially incomplete object localization, while RAMs expand the coverage to more coherent regions via affinity propagation.
Guided by the proposed region, FSL produces segmentation masks that accurately delineate the full extent of the referred object.
Notably, when attention maps activate only on salient parts (e.g., the blue shirt of the child in the first row), FSL is able to recover the complete object.
In more challenging cases, where attention maps highlight multiple similar instances or exhibit noisy activations (as exemplified in the last row), FSL correctly identifies the referred ``green van'' between the two candidates and suppresses spurious responses outside the ``dog'' region.
These examples corroborate the quantitative findings: attention provides a coarse spatial prior, while precise localization relies on semantic feature discrimination.

\begin{figure}[t]
    \centering
    \begin{minipage}{0.49\textwidth}
        \centering
        \includegraphics[width=\linewidth]{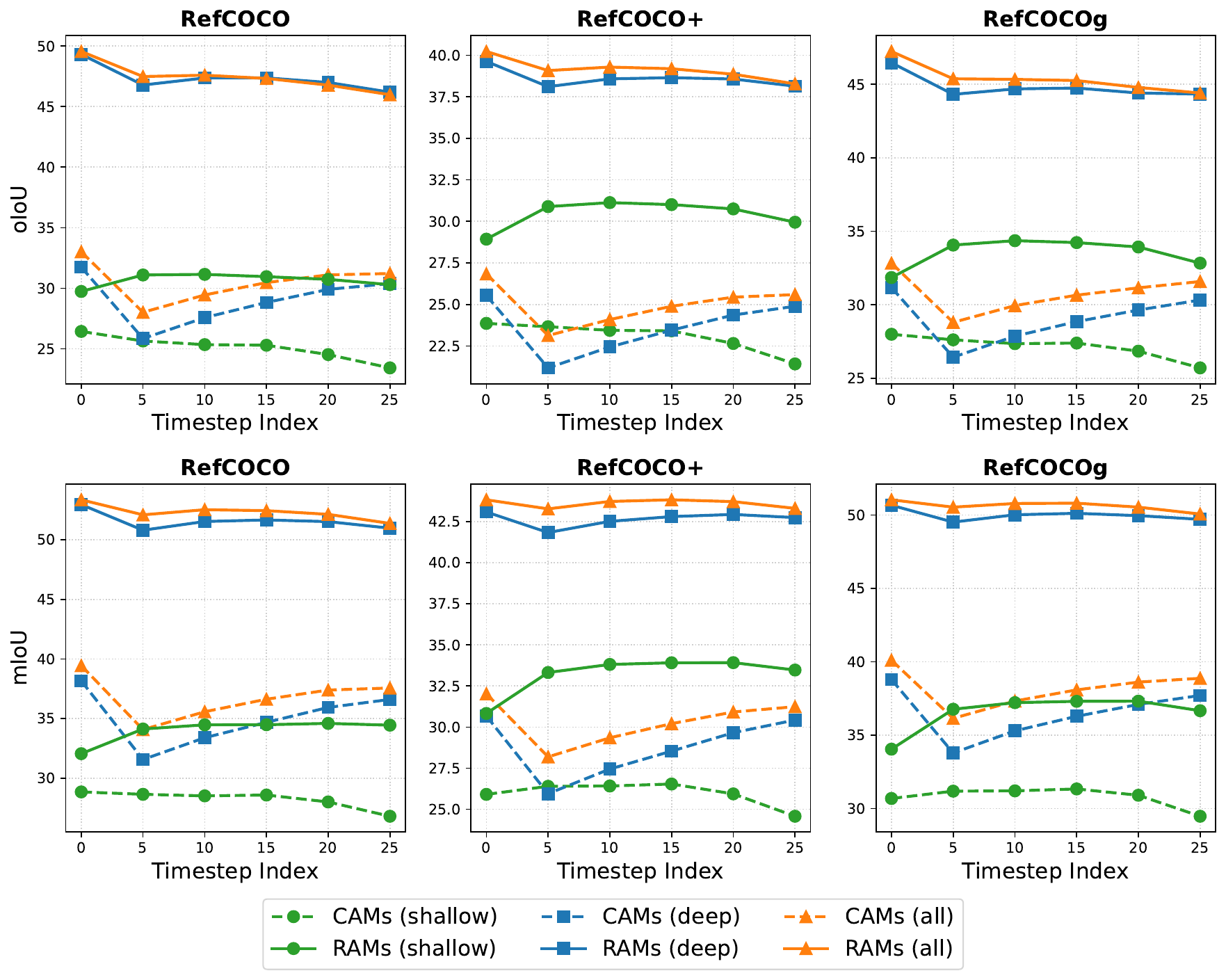}
        \caption{\textbf{Impact of timesteps and DiT blocks within the ARP module.} ``shallow'' and ``deep'' refer to accumulating attention weights over double-stream and single-stream DiT blocks, respectively, while ``all'' signifies attention accumulation over all DiT blocks.}
        \label{fig:ablation_attn}
    \end{minipage}
    \hfill
    \begin{minipage}{0.49\textwidth}
        \centering
        \includegraphics[width=\linewidth]{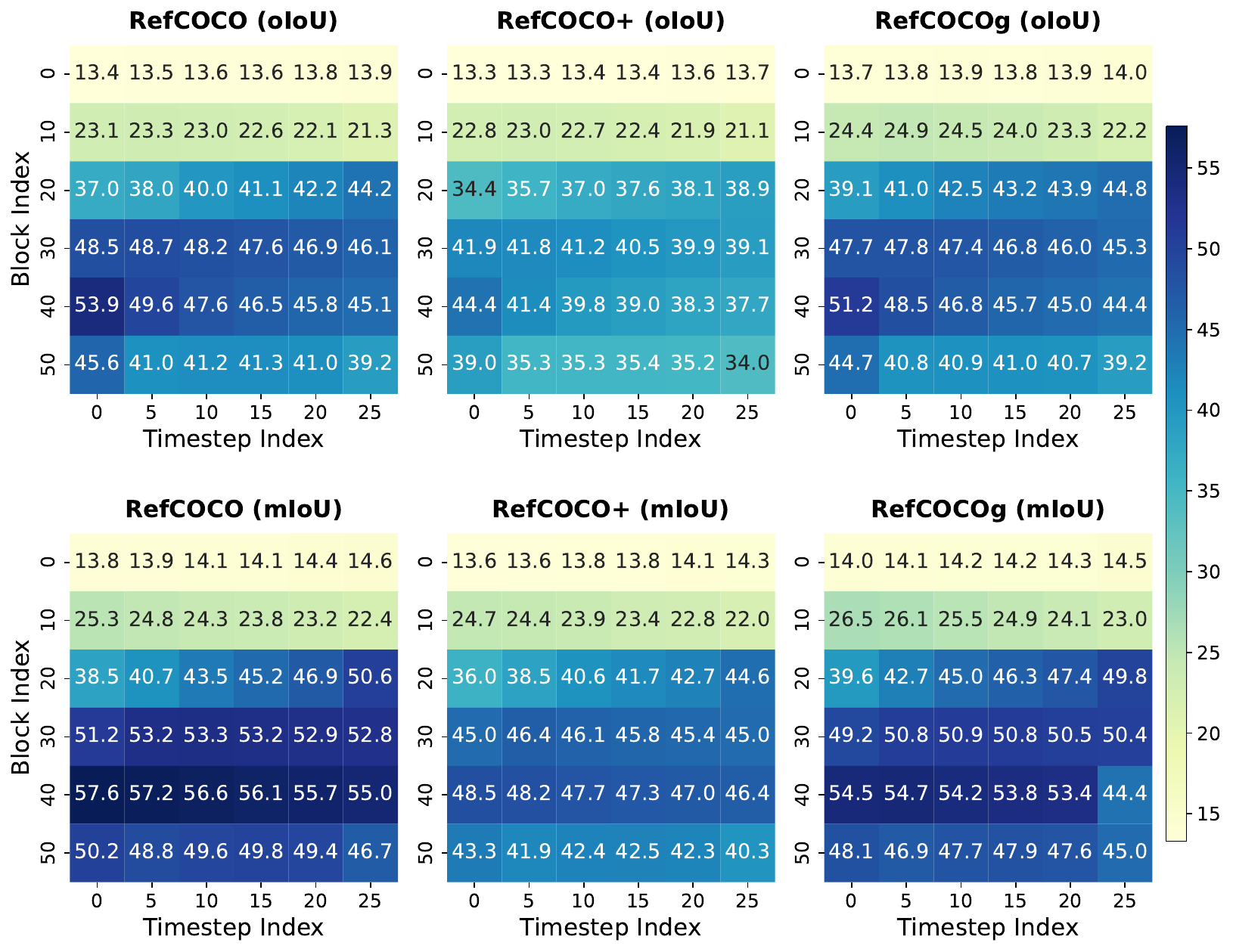}
        \caption{\textbf{Impact of timesteps and DiT blocks within the FSL module.} Heatmaps show oIoU (1st row) and mIoU (2nd row) on RefCOCO, RefCOCO+, and RefCOCOg datasets. Each cell represents segmentation performance when extracting features from a specific block at a given timestep.}
        \label{fig:ablation_feat}
    \end{minipage}
\end{figure}

\noindent \textbf{Impact of denoising timesteps and DiT blocks.}
For the ARP module, we examine the impact of denoising timesteps and DiT blocks on the quality of CAMs and RAMs.
Specifically, we evaluate oIoU and mIoU across denoising timesteps $t \in \{0, 5, 10, 15, 20, 25\}$, and analyze the contribution of different DiT components by accumulating attention weights over double-stream (i.e., shallow), single-stream (i.e., deep), and all blocks.
The experimental results illustrated in \cref{fig:ablation_attn} lead to several key observations.
First, RAMs (solid lines) consistently outperform CAMs (dotted lines) across all configurations, which corroborates the effectiveness of affinity propagation observed in \cref{tab:ablation_modules}.
Second, from an architectural perspective, aggregating attention from shallow blocks (green lines) yields substantially lower performance than using deep blocks (blue lines) or all blocks (orange lines).
Notably, accumulation over deep blocks achieves comparable performance with that over all blocks, indicating that semantic cues critical for referring object comprehension are mainly captured in deep layers.
Finally, we observe that the initial denoising step ($t=0$) yields peak performance across all benchmarks in terms of oIoU and mIoU.
This finding supports our design choice of single-step inference without a full iterative denoising process.

For the FSL module, we conduct experiments over timesteps $t^{\prime} \in \{0, 5, 10, 15, 20, 25\}$ and blocks $l^{\prime} \in \{0, 10, 20, 30, 40, 50\}$ to study the effect of feature extraction configurations.
As shown in \cref{fig:ablation_feat}, the combination of $t^{\prime}=0$ and $l^{\prime}=40$ consistently yields the peak segmentation performance across all benchmarks in terms of oIoU and mIoU.
These empirical findings resonate with our previous observations:
First, the initial timestep ($t^{\prime}=0$) already provides sufficiently discriminative features for referred object recognition, thereby eliminating the need for subsequent denoising iterations; Second, features extracted from deep layers (approximately two-thirds of the model depth) possess strong semantics that can be leveraged for accurate segmentation of the referred object.

\noindent \textbf{Ablation of feature extraction location.}
To determine an effective feature extraction location for the FSL module, we compare representations extracted from three locations within a DiT block: the outputs of Multi-Modal Self-Attention (MMSA), Adaptive Layer Normalization (AdaLN), and the block output.
As shown in \cref{tab:ablation_feat_extraction}, MMSA features consistently achieve the best performance across all benchmarks, outperforming both AdaLN and block features by a clear margin.
While prior work~\cite{gan2025unleashing} suggests that post-AdaLN features exhibit strong semantic correspondences, our results indicate that such features are less effective for referring image segmentation where precise cross-modal grounding is required.
We attribute the advantage of MMSA to its explicit modeling of interactions between textual and visual tokens, which yields more discriminative and semantically aligned representations for the referred object.

\noindent\begin{minipage}[t]{0.49\textwidth}
\vspace{0pt}
\noindent \textbf{Impact of thresholding.}
We study the effect of the threshold $\tau$ within the FSL module in \cref{fig:ablation_threshold}.
As $\tau$ increases from $0.1$ to $0.8$, both oIoU and mIoU scores increase steadily, suggesting that higher thresholds help emphasize more discriminative features and promote coherent clustering of semantically similar regions.
However, a slight performance degradation occurs at $\tau=0.9$.
This indicates that overly aggressive thresholding may discard useful semantic cues.
Based on this trade-off, we adopt $\tau=0.8$ for feature-based semantic localization.
\end{minipage}
\hfill
\begin{minipage}[t]{0.49\textwidth}
\vspace{0pt}
\centering
\includegraphics[width=\linewidth]{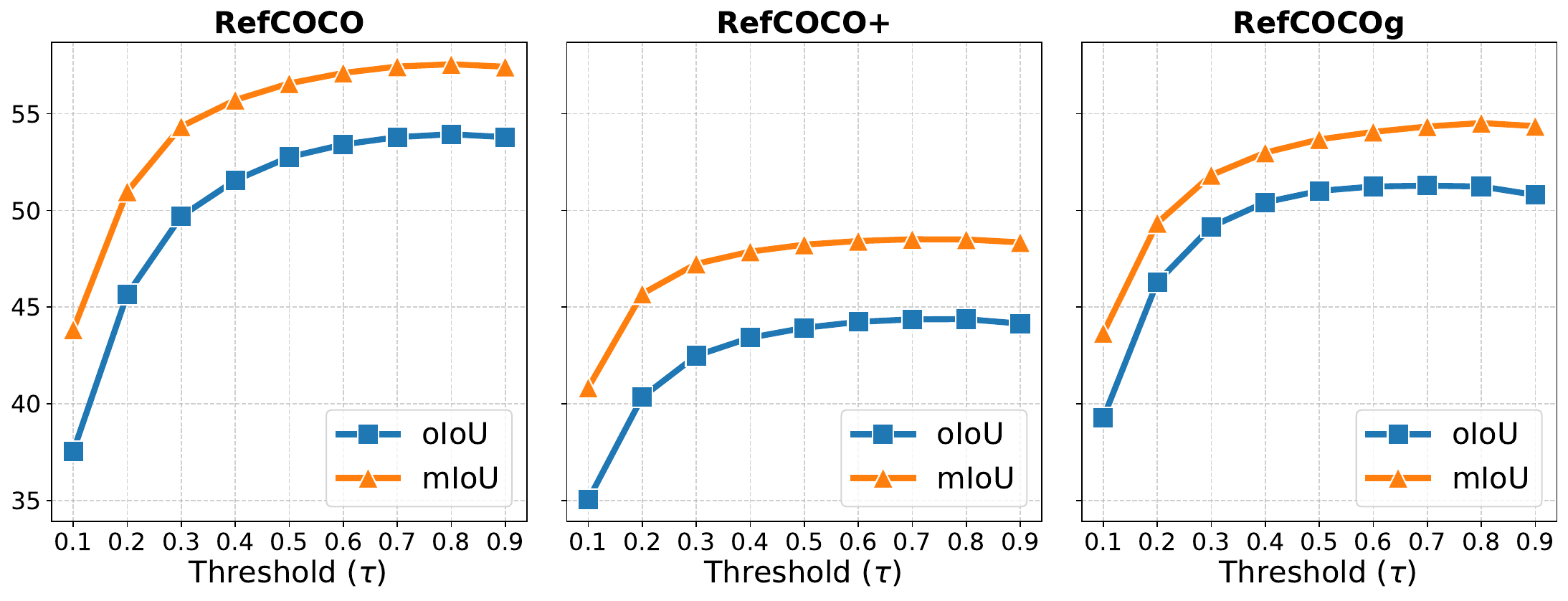}
\captionof{figure}{\textbf{Impact of thresholding within the FSL module}. The performance exhibits a steady increase until $\tau=0.8$, then a slight performance degradation occurs at $\tau=0.9$.}
\label{fig:ablation_threshold}
\end{minipage}

\begin{table}[t]
    \centering
    \small
    \caption{\textbf{Sensitivity to the instruction template.} We report changes in oIoU / mIoU on the \textit{val} splits relative to the default ``Remove'' instruction.}
    \label{tab:instruction_template}
    \begin{tabular}{ll|ccc}
    \toprule
    Model & Instruction & RefCOCO & RefCOCO+ & RefCOCOg \\
    \midrule
    \multirow{3}{*}{FLUX.2-klein (4B)}
        & Erase   & $+1.12 / +1.17$ & $+0.28 / +0.39$ & $+0.11 / +0.41$ \\
        & Delete  & $+0.97 / +0.77$ & $+0.53 / +0.50$ & $+0.10 / +0.29$ \\
        & Replace & $+0.27 / +1.18$ & $-1.13 / -0.75$ & $-1.09 / -0.26$ \\
    \midrule
    \multirow{3}{*}{Step1X-Edit (12B)}
        & Erase   & $-1.17 / -0.80$ & $-0.49 / -0.44$ & $-0.90 / -0.65$ \\
        & Delete  & $-1.53 / -1.00$ & $-1.03 / -0.90$ & $-0.94 / -0.71$ \\
        & Replace & $-4.72 / -3.81$ & $-3.60 / -3.72$ & $-5.36 / -4.54$ \\
    \bottomrule
    \end{tabular}
\end{table}

\noindent \textbf{Sensitivity to the instruction template.}
Since our framework converts each referring expression into a removal instruction, we examine how sensitive segmentation performance is to the choice of editing verb.
On the full \textit{val} splits, we replace ``Remove'' with ``Erase'', ``Delete'', or ``Replace'' while keeping the rest of each instruction unchanged.
As shown in \cref{tab:instruction_template}, the two removal synonyms, ``Erase'' and ``Delete'', yield modest changes of at most 1.53 oIoU points across both image editing models.
The effect of ``Replace'' varies more across models: performance changes by at most 1.13 oIoU points for FLUX.2-klein, whereas Step1X-Edit drops by 3.60-5.36 points.
One possible explanation is that removal encourages localization of the entire target, while replacement may emphasize only the regions that need modification.
Even with ``Replace'', Step1X-Edit achieves 49.22, 40.78, and 45.87 oIoU on RefCOCO, RefCOCO+, and RefCOCOg, respectively, outperforming RefAM's corresponding scores of 46.91, 38.57, and 45.53.
Overall, these results indicate limited sensitivity to the tested removal synonyms, but also show that segmentation performance depends partly on the operation specified and the underlying image editing model.

%% file: sections/5_conclusion.tex
\section{Conclusion}
We introduce a fundamentally different perspective on zero-shot referring image segmentation by reformulating it as an instruction-based image editing task.
Our analysis reveals that semantically separable representations already emerge at early denoising steps.
Building on this insight, we propose a simple yet effective training-free framework that combines attention-derived spatial priors with feature-level semantic separation to directly predict segmentation masks.
Extensive experiments demonstrate that our approach achieves state-of-the-art performance while remaining computationally efficient.
Our findings challenge the conventional view that object localization in RIS requires explicit spatial modeling, and instead suggest that it can arise naturally from the semantic grounding capability in generative models.
We hope this work motivates further exploration of generative representations as a unified foundation for both synthesis and discriminative vision-language tasks.

%% file: sections/x_appendix.tex
\begin{center}
    {\Large \bf Appendix}
\end{center}

\section{Feature Separability: Definition, Rationale, and Implementation}
\label{app:feature_separability}

\noindent \textbf{Definition.}
Let $\mathbf{F}^{(t,l)} \in \mathbb{R}^{N_v \times D}$ denote MMSA output features of clean image tokens within the $l$-th DiT block at timestep $t$.
Given the ground-truth binary mask, we denote the sets of foreground and background token indices as $\Omega_{\text{fg}}$ and $\Omega_{\text{bg}}$, respectively.
The class-wise means are defined as:
\begin{equation}
\begin{aligned}
    \mu_{\text{fg}}^{(t,l)} &= \frac{1}{|\Omega_{\text{fg}}|} \sum_{i \in \Omega_{\text{fg}}} \mathbf{F}_i^{(t,l)}, \\
    \mu_{\text{bg}}^{(t,l)} &= \frac{1}{|\Omega_{\text{bg}}|} \sum_{i \in \Omega_{\text{bg}}} \mathbf{F}_i^{(t,l)}.
\end{aligned}
\end{equation}
The corresponding within-class variances are defined as:
\begin{equation}
\begin{aligned}
    \sigma_{\text{fg}}^{2(t,l)} &= \frac{1}{|\Omega_{\text{fg}}|} \sum_{i \in \Omega_{\text{fg}}} \left\| \mathbf{F}_i^{(t,l)} - \mu_{\text{fg}}^{(t,l)} \right\|^2, \\
    \sigma_{\text{bg}}^{2(t,l)} &= \frac{1}{|\Omega_{\text{bg}}|} \sum_{i \in \Omega_{\text{bg}}} \left\| \mathbf{F}_i^{(t,l)} - \mu_{\text{bg}}^{(t,l)} \right\|^2.
\end{aligned}
\end{equation}
The feature separability is formulated as:
\begin{equation}
\mathcal{J}^{(t,l)} = \frac{\left\| \mu_{\text{fg}}^{(t,l)} - \mu_{\text{bg}}^{(t,l)} \right\|^2}{\sigma_{\text{fg}}^{2(t,l)} + \sigma_{\text{bg}}^{2(t,l)}}.
\end{equation}

\noindent \textbf{Connection to Fisher Discriminant.}
The metric $\mathcal{J}^{(t,l)}$ quantifies the ratio of inter-class separation to intra-class variance, evaluating how effectively foreground and background features are disentangled in the representation space.
This formulation is closely related to the classical Fisher discriminant criterion for two classes in multivariate space:
\begin{equation}
    \mathcal{J} = (\mu_{\text{fg}} - \mu_{\text{bg}})^\top (\Sigma_{\text{fg}} + \Sigma_{\text{bg}})^{-1} (\mu_{\text{fg}} - \mu_{\text{bg}}),
\end{equation}
where $\Sigma_{\text{fg}}$ and $\Sigma_{\text{bg}}$ denote the respective covariance matrices.
Our formulation $\mathcal{J}^{(t,l)}$ can be viewed as a simplified isotropic variant.
By approximating the covariance matrices as isotropic (i.e., $\Sigma \approx \sigma^{2(t,l)} \mathbf{I}$), the inverse covariance simplifies to a scalar division.
This yields a tractable and numerically stable measure that is highly suitable for high-dimensional features.

\noindent \textbf{Implementation.}
We evaluate the feature separability metric $\mathcal{J}^{(t,l)}$ on FLUX.2-klein~\cite{flux-2-2025} and Step1X-Edit~\cite{liu2025step1x} across six uniformly sampled timesteps.
For both models, we extract features from the DiT block located at approximately two-thirds of the model depth ($l=20$ of 32 for FLUX.2-klein and $l=40$ of 57 for Step1X-Edit), where semantic representations are typically most pronounced.
To obtain the spatial location sets $\Omega_{\text{fg}}$ and $\Omega_{\text{bg}}$, we resize ground-truth segmentation masks to the spatial resolution of clean image tokens.
All feature vectors are $\ell_2$-normalized prior to computing $\mathcal{J}^{(t,l)}$ to ensure scale consistency across timesteps.
For the boxplots in \cref{fig:overview}, we conduct this evaluation on the RefCOCO~\cite{nagaraja2016modeling} \textit{val} split.

\begin{table}[t]
    \centering
    \small
    \caption{\textbf{Quantitative feature separability across the denoising trajectory.} We report the median of $\mathcal{J}^{(t,l)}$ over 500 randomly sampled examples from each \textit{val} split. The maximum of each row is in \textbf{bold}.}
    \label{tab:separability}
    \begin{tabular}{ll|cccccc}
    \toprule
    Model & Dataset & 1\% & 20\% & 40\% & 60\% & 80\% & 100\% \\
    \midrule
    \multirow{3}{*}{Step1X-Edit (12B)} & RefCOCO  & \textbf{0.403} & 0.324 & 0.358 & 0.361 & 0.366 & 0.361 \\
                                       & RefCOCO+ & \textbf{0.286} & 0.212 & 0.208 & 0.206 & 0.200 & 0.182 \\
                                       & RefCOCOg & \textbf{0.407} & 0.330 & 0.343 & 0.358 & 0.372 & 0.384 \\
    \midrule
    \multirow{3}{*}{FLUX.2-klein (9B)} & RefCOCO  & 0.318 & 0.296 & 0.305 & 0.322 & 0.342 & \textbf{0.343} \\
                                       & RefCOCO+ & 0.234 & 0.232 & 0.239 & 0.251 & \textbf{0.257} & 0.247 \\
                                       & RefCOCOg & 0.301 & 0.294 & 0.299 & 0.314 & \textbf{0.318} & 0.293 \\
    \bottomrule
    \end{tabular}
\end{table}

\noindent \textbf{Quantitative results across benchmarks.}
To quantify this trend beyond RefCOCO, we compute $\mathcal{J}^{(t,l)}$ on 500 randomly sampled examples from each \textit{val} split of RefCOCO, RefCOCO+, and RefCOCOg, and report the median at six stages of the denoising trajectory in \cref{tab:separability}.
For Step1X-Edit, the median separability is highest at the earliest evaluated stage on all three datasets.
For FLUX.2-klein, the values at the earliest stage already reach 92.7\%, 91.1\%, and 94.7\% of the maximum observed values on RefCOCO, RefCOCO+, and RefCOCOg, respectively.
Thus, although FLUX.2-klein sometimes reaches its maximum later in the trajectory, a strong foreground-background discriminative signal is already present at the earliest evaluated timestep.

\section{Inference Efficiency Analysis}
\label{app:inference_efficiency}

\begin{table}[t]
    \centering
    \small
    \caption{\textbf{Inference efficiency of the DiT backbone in our framework.} Latency and peak GPU memory are measured on a single NVIDIA A100 GPU with a batch size of 1. The size level follows the officially recommended settings.}
    \label{tab:efficiency}
    \begin{tabular}{lcccc}
    \toprule
    Model & Size Level & Param. & Latency (ms) & Memory (GB) \\
    \midrule
    FLUX.2-klein & $1024$ & 9B  & 1493.02 & 22.27 \\
    Step1X-Edit  & $512$   & 12B & 603.78  & 25.31 \\
    \bottomrule
    \end{tabular}
\end{table}

\begin{table}[t]
    \centering
    \small
    \caption{\textbf{End-to-end efficiency comparison with baselines.} All methods are evaluated on the same NVIDIA A100 GPU with 5 warm-up runs, averaging over 500 samples from the RefCOCO \textit{val} split.}
    \label{tab:efficiency_baselines}
    \begin{tabular}{lcc}
    \toprule
    Method & End-to-end latency (ms) & Peak memory (GB) \\
    \midrule
    Global-Local (SAM + CLIP)~\cite{yu2023zero} & 1203 & 3.3 \\
    HybridGL~\cite{liu2025hybrid}                & 942  & 6.9 \\
    RefAM (FLUX.1-dev)~\cite{kukleva2025refam}    & 2367 & 37.0 \\
    Ours (Step1X-Edit)                           & \textbf{853} & 40.8 \\
    \bottomrule
    \end{tabular}
\end{table}

We evaluate the inference efficiency of our framework in terms of inference latency and memory usage.
All measurements are conducted on a single NVIDIA A100 GPU with a batch size of 1.
Following the officially recommended settings, we use a size level of $512$ for Step1X-Edit and $1024$ for FLUX.2-klein.
As shown in \cref{tab:efficiency}, the DiT backbone of Step1X-Edit achieves an average latency of 603.78 ms with a peak memory usage of 25.31 GB, while FLUX.2-klein requires 1493.02 ms and 22.27 GB.
The higher latency of FLUX.2-klein is mainly due to its larger input resolution.

We further compare the complete pipeline with representative baselines in \cref{tab:efficiency_baselines}.
Note that \cref{tab:efficiency} reports the DiT backbone only, whereas \cref{tab:efficiency_baselines} measures end-to-end inference.
Owing to single-step inference, our method achieves the lowest end-to-end latency, running $1.1\times$, $1.4\times$, and $2.8\times$ faster than HybridGL, Global-Local, and RefAM, respectively.
However, the 12B configuration reaches a peak memory of 40.8\,GB, most of which is attributable to the underlying editor: a Step1X-Edit forward pass alone requires 39.4\,GB.

\section{Impact of Editing Instruction Refinement}
\label{sec:instruction_refinement}

\begin{table}[tbp]
    \centering
    \small
    \caption{\textbf{Impact of editing instruction refinement}. We compare our method against Global-Local~\cite{yu2023zero} and HybridGL~\cite{liu2025hybrid} using both the original and refined referring expressions.}
    \label{tab:ablation_expression}
    \begin{tabular}{c|l|cc|cc|cc}
    \toprule
    \multirow{2}{*}{Refinement}     & \multirow{2}{*}{Method} & \multicolumn{2}{c|}{RefCOCO} & \multicolumn{2}{c|}{RefCOCO+} & \multicolumn{2}{c}{RefCOCOg} \\
                                    & & oIoU             & mIoU & oIoU & mIoU & oIoU & mIoU \\
    \midrule
    \multirow{3}{*}{$\times$}       & Global-Local       & 24.55 & 31.83 & 26.62 & 34.97 & 28.92 & 40.66 \\
                                    & HybridGL           & 41.81 & 49.48 & 35.74 & 43.40 & 42.47 & 51.25 \\
                                    & Ours (Step1X-Edit) & \textbf{51.91} & \textbf{55.79} & \textbf{41.06} & \textbf{44.39} & \textbf{48.92} & \textbf{52.43} \\
    \midrule
    \multirow{3}{*}{\checkmark}     & Global-Local       & 24.58 & 33.44 & 27.14 & 36.69 & 29.36 & 41.54 \\
                                    & HybridGL           & 44.14 & 52.55 & 38.53 & 46.78 & 43.74 & 52.78 \\
                                    & Ours (Step1X-Edit) & \textbf{53.94} & \textbf{57.56} & \textbf{44.38} & \textbf{48.50} & \textbf{51.23} & \textbf{54.52} \\
    \bottomrule
    \end{tabular}
\end{table}

We investigate the impact of editing instruction refinement on segmentation performance to understand how instruction clarity influences the model's internal priors.
Each referring expression is first converted to ``Remove \{expression\}'' and then processed together with the source image by the built-in refinement pipeline of Step1X-Edit~\cite{yin2025reasonedit}.
The refiner is instructed to preserve the original meaning, retain instructions that are already clear, and otherwise produce a concrete and actionable editing command.
For a fair comparison, we also evaluate Global-Local~\cite{yu2023zero} and the recent state-of-the-art RIS method HybridGL~\cite{liu2025hybrid} using refined expressions with only the editing directive (e.g., ``remove'') discarded, since these methods take referring expressions rather than editing commands.
As shown in \cref{tab:ablation_expression}, refinement generally improves all methods.
We attribute this improvement to improved distributional alignment, as the refined prompts more closely match the data distribution seen during training of the image editing model.
Although the baselines also benefit from clearer prompts, our method maintains a substantial performance margin.
Without refinement, our method outperforms HybridGL by 10.10, 5.32, and 6.45 oIoU on RefCOCO, RefCOCO+, and RefCOCOg, respectively, and under the same refinement pipeline the margins remain 9.80, 5.85, and 7.49 oIoU.
Thus, instruction refinement is beneficial but does not explain the advantage of our method.

\section{Qualitative Results of SAM 3}
\label{app:sam3}

\begin{figure}[t]
    \centering
    \includegraphics[width=0.8\linewidth]{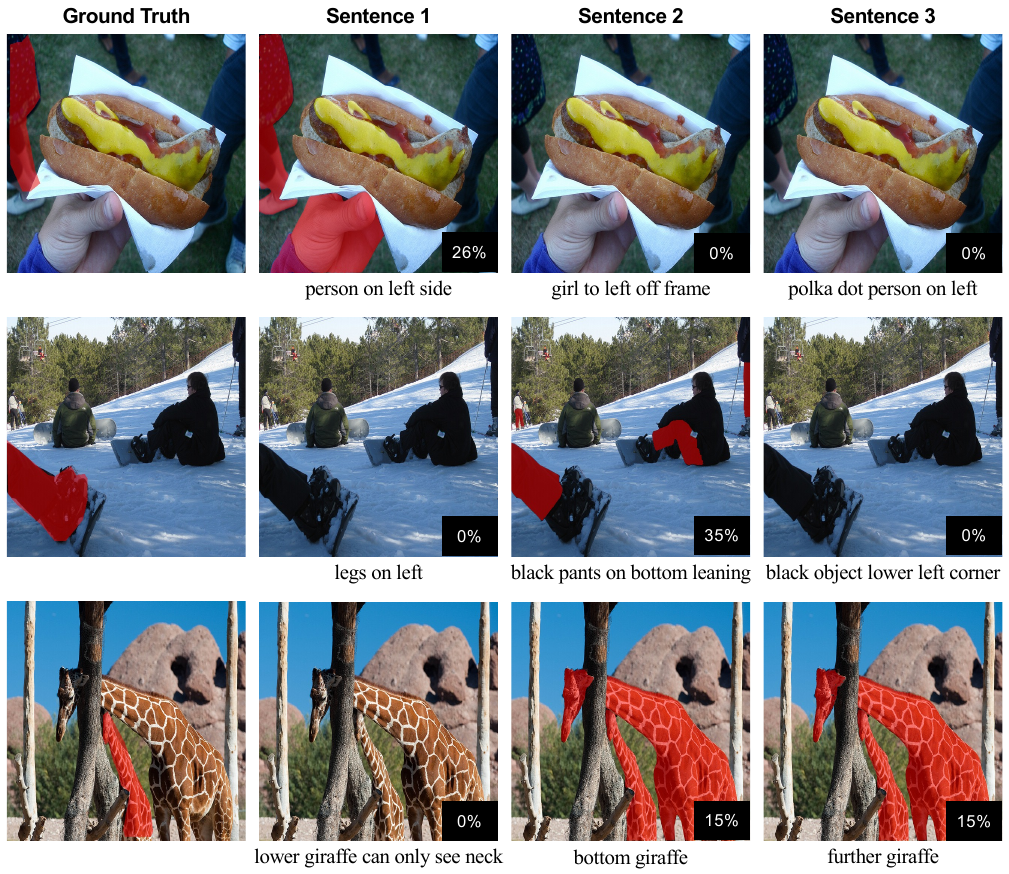}
    \caption{\textbf{Qualitative results of SAM 3}. The referring expression is shown at the bottom of each image. The corresponding IoU score is displayed at the bottom-right corner in each image.}
    \label{fig:sam3}
\end{figure}

The Segment Anything Model (SAM) 3~\cite{carion2025sam} takes concept prompts to retrieve segmentation masks for all matching instances in an image.
While SAM 3 demonstrates strong capability in producing high-quality object boundaries, we observe that it exhibits two fundamental limitations when applied to referring image segmentation.
First, SAM 3 often fails to predict the presence of the target object conditioned on the referring expression, resulting in a low recall rate.
Second, even when the target object is successfully detected, the model tends to segment all instances belonging to the same semantic category, without adhering to instance-level constraints such as spatial relations specified in the referring expression.
We present illustrative examples in \cref{fig:sam3} to demonstrate these failure modes.
As can be seen, SAM 3 frequently misses the target object entirely, resulting in a high proportion of zero-IoU predictions.
In the last example, even when the prompt specifies the position of the ``giraffe'', the model still produces masks covering all giraffes in the scene.

\section{Limitations and Discussions}
\label{app:limitations}

\begin{figure}[t]
    \centering
    \includegraphics[width=\linewidth]{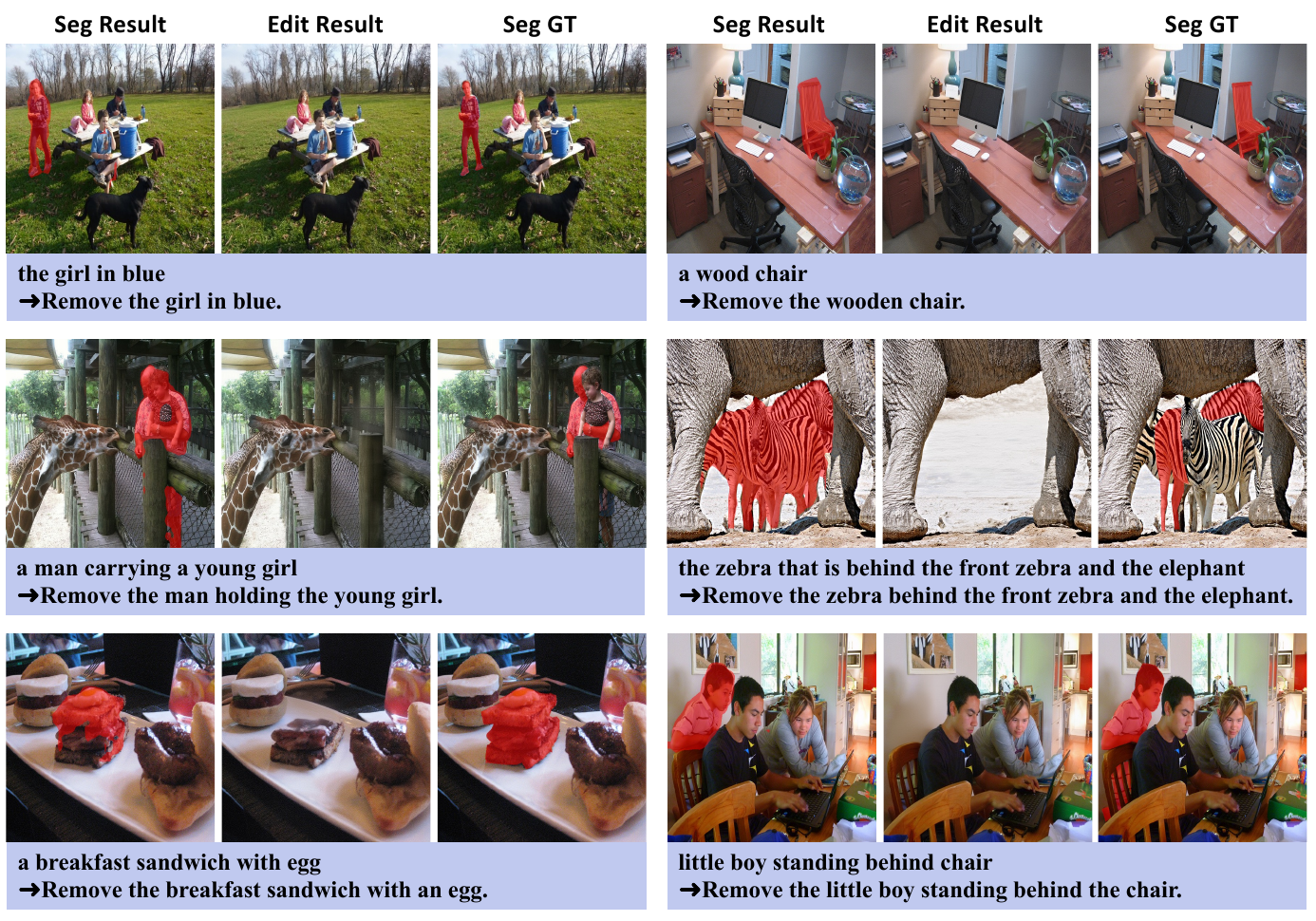}
    \caption{\textbf{Examples of failure cases}. The limitations primarily arise from the inherent capabilities of the underlying image editing model, as the predicted masks closely follow the edited regions.}
    \label{fig:failure_cases}
\end{figure}

We identify three main types of failure cases, as illustrated in \cref{fig:failure_cases}.
The examples in the first row show that the image editing model struggles with fine-grained object boundaries, leading to imprecise segmentation along detailed contours.
The second row demonstrates cases where the model fails to distinguish the referred instance and instead activates on all objects of the same category (e.g., all people or all zebras in the image).
The third row highlights situations in which the model cannot capture the complete extent of the target object, resulting in partial segmentation.
Importantly, \emph{these limitations primarily stem from the inherent capabilities of the underlying image editing model rather than our framework design}, as the predicted masks remain consistent with the edited regions.

\begin{table}[t]
    \centering
    \small
    \caption{\textbf{Quantitative failure-case analysis.} We report the ratio of each category among 1,000 randomly sampled \textit{val} expressions per dataset and the corresponding oIoU change (Ratio / $\Delta$oIoU). Categories are not mutually exclusive.}
    \label{tab:failure_cases}
    \begin{tabular}{l|ccc}
    \toprule
    Failure case & RefCOCO & RefCOCO+ & RefCOCOg \\
    \midrule
    Overlapping instances & $37.6\% / -02.54$ & $40.5\% / -01.85$ & $27.6\% / -08.52$ \\
    Spatial relations     & $67.7\% / -04.18$ & $17.2\% / -06.61$ & $40.5\% / -06.32$ \\
    Thin edges            & $13.2\% / -16.33$ & $13.0\% / -11.49$ & $21.4\% / -15.98$ \\
    Occlusions            & $11.0\% / -12.59$ & $12.7\% / -03.26$ & $15.5\% / -11.67$ \\
    Partial objects       & $25.3\% / -02.53$ & $24.6\% / -02.72$ & $24.3\% / -02.25$ \\
    \bottomrule
    \end{tabular}
\end{table}

To quantify these failure modes, we randomly sample 1,000 \textit{val} expressions from each dataset and group them using the referring expressions and ground-truth masks:
(1) \textit{overlapping instances}, where the target spatially overlaps another instance of the same category;
(2) \textit{spatial relations}, where the target requires relational cues;
(3) \textit{thin edges}, where the target contains narrow structures that span very few spatial tokens in the latent space;
(4) \textit{occlusions}, where the visible target is separated into disconnected regions or has substantial missing regions in its silhouette; and
(5) \textit{partial objects}, where the target is truncated by the image boundary or explicitly described as only part of an object.
As shown in \cref{tab:failure_cases}, the largest and most consistent degradation occurs for thin edges, which we attribute primarily to the limited spatial resolution of the editor's latent features.
One possible extension is to use our prediction as a semantic prompt for SAM to recover finer boundaries; we leave this to future work, as the present framework is intentionally designed to remain segmenter-free.

\section{Broader Impact}
\label{app:broader_impact}
This work advances zero-shot referring image segmentation by leveraging the implicit localization capability of instruction-based image editing models.
Such capability may benefit applications in visual understanding where language-driven object grounding is required.
On the other hand, the underlying image editing models repurposed in this work are capable of synthesizing realistic image manipulations, which raises concerns about potential misuse for generating misleading or harmful visual content.
However, our framework exclusively extracts internal representations from these models for localization purposes and does not involve any image generation or modification.
We therefore do not introduce additional risks beyond those already associated with the pretrained models themselves.